\documentclass[a4paper,fleqn]{cas-sc}

\usepackage[authoryear,longnamesfirst]{natbib}
\usepackage{amsmath,amsfonts}
\usepackage{algorithmic}
\usepackage{algorithm}
\usepackage{array}
\usepackage[caption=false,font=normalsize,labelfont=sf,textfont=sf]{subfig}
\usepackage{textcomp}
\usepackage{stfloats}
\usepackage{url}
\usepackage{verbatim}
\usepackage{graphicx}
\usepackage{multirow}
\usepackage{booktabs}
\usepackage{balance}
\usepackage{cite}
\usepackage{lineno,hyperref}
\usepackage{graphicx}
\usepackage{enumitem} 
\setlist[itemize]{leftmargin=1cm} 
\def\tsc#1{\csdef{#1}{\textsc{\lowercase{#1}}\xspace}}
\tsc{WGM}
\tsc{QE}
\tsc{EP}
\tsc{PMS}
\tsc{BEC}
\tsc{DE}

\begin{document}
\let\WriteBookmarks\relax
\def\floatpagepagefraction{1}
\def\textpagefraction{.001}
\shorttitle{Federated Class-Incremental Learning with Prompting}
\shortauthors{Xin Luo et~al.}

\title [mode = title]{Federated Class-Incremental Learning with Prompting}                      

\newcommand{\orcidauthorA}{0000-0002-6901-5476}
\newcommand{\orcidauthorB}{0009-0002-4210-9904}
\newcommand{\orcidauthorC}{0009-0006-2904-3508}
\newcommand{\orcidauthorD}{0000-0002-5822-5646}
\newcommand{\orcidauthorE}{0000-0002-3481-4892}
\newcommand{\orcidauthorF}{0000-0001-9972-7370}

\affiliation[1]{organization={School of Software, Shandong University},
                city={Jinan},
                postcode={250101}, 
                country={China}}
\affiliation[2]{organization={Department of Computer Science and Technology, Tsinghua University},
                city={Beijing},
                postcode={100084}, 
                country={China}}

\author[1]{Xin Luo}[
    orcid=\orcidauthorA
]

\author[1]{Fang-Yi Liang}[
    orcid=\orcidauthorB
]
\author[1]{Jiale Liu}[
    orcid=\orcidauthorC
]
\cormark[1]
\ead{carreliu@hotmail.com}  

\author[2]{Yu-Wei Zhan}[
    orcid=\orcidauthorD
]
\cormark[1]
\ead{zhanyuweilif@gmail.com}  

\author[1]{Zhen-Duo Chen}[
    orcid=\orcidauthorE
]
\author[1]{Xin-Shun Xu}[
    orcid=\orcidauthorF,
]

\cortext[1]{
    Corresponding author.
}

\begin{abstract}
As Web technology continues to develop, it has become common to use data stored on different clients. At the same time, federated learning has received attention due to its ability to protect data privacy when letting models learn from data distributed across various clients. However, most existing works assume that the client's data are fixed. In real-world scenarios, such an assumption is often not true as data may be continuously generated and new classes may also appear. To this end, we focus on the practical and challenging federated class-incremental learning (FCIL) problem. For FCIL, the local and global models may suffer from catastrophic forgetting on old classes caused by the arrival of new classes, and the data distributions of clients are non-independent and identically distributed (non-iid).

In this paper, we propose a method called Federated Class-Incremental Learning with PrompTing (FCILPT). Given privacy and memory constraints, FCILPT does not use rehearsal buffers for old data. We use prompts to ease catastrophic forgetting of old classes. Specifically, we encode task-relevant and task-irrelevant knowledge into prompts, preserving old and new knowledge of local clients and solving catastrophic forgetting. We first sort task information in the prompt pool on local clients to align task information across clients before global aggregation. It ensures that the same task's knowledge is fully integrated, addressing the non-iid problem caused by class imbalance across clients under the same incremental task. Experiments on CIFAR-100, ImageNet-Subset, and TinyImageNet show that FCILPT achieves significant accuracy improvements over state-of-the-art methods.
\end{abstract}




\maketitle

\section{Introduction}
With the rapid development of Web, communication has become very convenient. Currently, data is widely generated and stored, and distributedly storing data is gradually becoming mainstream. Such a situation results in the need to transmit data to form a large training set to learn machine learning models. However, during this procedure, there may be many potential problems in data transmission, such as data privacy issues.

To not only using distributed data to train models but also providing secure privacy protection for data, federated learning (FL) \citep{mcmahan2017communication,li2020federated,tan2022fedproto,ma2022layer} has been proposed and widely investigated in recent years due to its ability to securely train models without compromising data privacy. A major problem faced by federated learning is that the clients' data are non-independent and identically distributed (non-iid) \citep{mcmahan2017communication,sattler2020clustered}. Directly aggregating local models, which are trained in clients with each client's own data, will lead to the performance degradation of the global model. Existing approaches \citep{zhang2022fine,tan2022fedproto,ma2022layer} have tried to  solve the non-iid problem by different techniques, such as distillation, prototype, personalized federated learning and so on. Although federated learning has achieved great success, most existing FL methods assume that each local client has fixed static data. Such assumption may fail in many real-world cases as local clients may continuously generate or receive new data. Furthermore, new data may carry new classes which local clients did not know. In such situations, existing FL methods have to retrain models from scratch. As training machine learning models may cause huge costs, it is  unsustainable and unaffordable to retrain models once new data with new classes is shown. 

In literature, if the setting of federated learning is not considered, the class-incremental learning could avoid the huge costs of retraining by incrementally learn models \citep{rebuffi2017icarl,aljundi2018memory,aljundi2019gradient,wu2022class,agarwal2022semantics} and has shown its effectiveness. Existing class-incremental learning methods can be roughly divided into four categories, namely architecture-based methods \citep{wu2022class}, regularization-based methods \citep{aljundi2018memory}, rehearsal-based methods \citep{rebuffi2017icarl,aljundi2018memory}, and generative-based methods \citep{agarwal2022semantics}. However, they are not designed for federated learning setting. Besides, existing methods also have some shortcomings. For example, rehearsal-based methods, which are more commonly used than other kinds of methods, are easy to implement and usually have high performance. They keep some exemplars of old data in the memory buffer, and when the new task arrives, exemplars in the buffer are added to the new task to train the model together. Such design may fail to work well with federated learning. First, as the number of dynamically arriving tasks keeps rising, we have to make a trade-off between keeping more old tasks and keeping more exemplars of old data due to memory constraints \citep{dong2022federated}.  Second, federated learning has extremely high security requirements\citep{shokri2015privacy}, so the use of buffers may not be allowed. 

Recently, above challenge has been noted and is defined as the federated class-incremental learning (FCIL) problem. A recent approach \citep{dong2022federated} takes a tentative look at incremental learning in federated scenarios. However, this method still uses memory buffer, and proposes a task transition detection mechanism based on entropy function. When the arrival of a new task is detected, the method adds the exemplars of the old tasks to the buffer. Since the buffer size is fixed, when the number of tasks increases, the exemplars of each old task keep decreasing, and the performance can not be guaranteed. Even with distillation on top of the memory buffer, the performance of the server-side aggregation model is still far behind that of centralized training, so we can see that the catastrophic forgetting problem \citep{mccloskey1989catastrophic} in class-incremental learning is not solved.

In this paper, we focus on the federated class-incremental learning (FCIL) problem, where the local and global models suffer from catastrophic forgetting on old classes caused by the arrival of new classes and the data distributions of clients are non-iid. Catastrophic forgetting \citep{mccloskey1989catastrophic} is caused by non-stationary streaming data on the local client. In addition, due to the problem that the data distributions of different clients for the current task are non-iid, the performance of the aggregation model will further decrease. Considering that rehearsal-based methods have the problems of limited memory, privacy insecurity and poor performance in federated learning scenarios, we decided to use a rehearsal-free method to alleviate the catastrophic forgetting of streaming tasks on local clients and non-iid problem among different clients under the same task. Taking inspiration from the field of natural language processing (NLP), we encode each task-relevant knowledge into learnable prompts \citep{lester2021power,li2021prefix,wang2022learning} and keep the pre-trained vision transformer (ViT) \citep{dosovitskiyimage} frozen. Referring to the approach of L2P \citep{wang2022learning}, we extend prompts into prompts pools, adding learnable prompt keys sections. With the prompt keys, we can implement an instance-wise prompt query mechanism, which can accurately select task-relevant prompts for an instance without knowing the current task identity. Specifically, we encode task-relevant information into the prompt pool, including task-specific information and task-similar information. Task-specific information contains knowledge unique to the current task, while task-similar information contains consistent knowledge shared by multiple similar tasks. Task-irrelevant information encodes all the tasks that have arrived, and obtains the possible common information between all the tasks. These three types of prompts preserve all the old and new knowledge of the local clients and solve the catastrophic forgetting. Before aggregating on the server, we first sort the prompt pool on the local client to align the task information from different clients. The sorted prompt pools from clients are aggregated to ensure that the three types of prompt information for the current task on different clients are fully integrated, and the problem of non-iid caused by the lack of classes of different clients under the same task is solved. To summarize, the main contributions of this paper are as follows:
\begin{itemize}
	\item A new and novel method FCILPT is proposed for the practical and challenging federated class-incremental learning (FCIL) problem. To the best of our knowledge, we are the first to apply prompt learning to FCIL.  To promote the development of this field, the code of our method is already available at \href{https://anonymous.4open.science/r/FCILPT-DAB}{Anonymous GitHub}.
	\item Three types of prompt encoding information are proposed, including task-specific prompt, task-similar prompt, and task-irrelevant prompt, which can save the knowledge of all tasks seen so far in the form of small learnable parameters to solve the catastrophic forgetting of the local clients.
	\item A variety of information related to tasks in the local clients' prompt pool is sorted, and the task information of different clients is aligned and fully integrated, so as to solve the non-iid problem caused by the lack of classes in different clients under the same task.
	\item Extensive experiments have been conducted on three benchmark datasets. And the experimental results show that FCILPT significantly outperforms the state-of-the-art methods.
\end{itemize}

\section{Related Work}
\subsection{Class-Incremental Learning}
Class-incremental learning (CIL) retains knowledge of old classes while learning new classes arriving in streams. In class-incremental learning, the task identity is unknown, which brings great difficulties to the learning of the model. The existing methods for solving class-incremental learning could be roughly divided into four categories, i.e., architecture-based methods, regularization-based ones, rehearsal-based ones, and generative-based ones.

Architecture-based methods \citep{fernando2017pathnet,golkar2019continual,hung2019compacting,rusu2016progressive,aljundi2019gradient,Douillard2022DyTox,zhou2024expandable,nguyen2024class,gao2024apm} dynamically create new independent modules in response to new tasks that arrive continuously. For example, DyTox \citep{Douillard2022DyTox} leverages the Transformer architecture and dynamically expands task-specific tokens to adapt to new tasks while keeping most parameters shared across all tasks. Specifically, DyTox uses a novel Task-Attention Block (TAB) that takes task-specific tokens and patch tokens as input, and produces task-specialized embeddings through a task-attention layer. However, these methods may have two  drawbacks. One is the high local overhead of the created module parameter size when the number of tasks increases. The other one is that old modules will inevitably make noise on the representations of new classes, leading to performance degradation on them.

Regularization-based methods \citep{aljundi2018memory,kirkpatrick2017overcoming,li2017learning,zenke2017continual} constraint the parameters with higher importance of the previous tasks. Limiting updates or penalizing the parameters of the network could retains the knowledge of the old tasks when new task data coming. For example, Elastic Weight Consolidation (EWC) \citep{kirkpatrick2017overcoming} and online EWC (oEWC) \citep{schwarz2018progress} compute synaptic importance using the diagonal Fisher information matrix as an approximation of the Hessian. Attention attractor network \citep{ren2019incremental} uses old weights to train a set of new weights that can recognize new classes, ensuring that old knowledge is not forgotten. It is found that regularization-based methods may perform worse than other types of class-incremental methods. For instance, in the case of complex datasets, the performance degradation of the regularization-based methods is large \citep{wu2019large}.

Rehearsal-based methods \citep{rebuffi2017icarl,aljundi2019gradient} build a memory buffer in which partial representative part of data of each old tasks are selected for keeping, which is called exemplars. When a new task arrives, the exemplars are combined with new data to jointly train the model. 
A large number of studies have found that how to select exemplars from old data plays a decisive role in the performance of the model. For example, iCaRL \citep{rebuffi2017icarl} selects the exemplars in a greedy manner under cardinality constraints by approximating the mean of the training data. GSS \citep{aljundi2019gradient} introduces gradient-based sampling to store the best selected exemplars in the memory buffer. However, in the case of high privacy security requirements, the memory buffer is not allowed to be used. In this paper, to match the privacy security scenario of federated learning, we do not utilize memory buffer.

Different from retaining the original old data, generative-based methods \citep{van2020brain,agarwal2022semantics,cao2024generative} do not need to keep exemplars of old data, but strengthen the memory of old knowledge by generating pseudo old data. These methods usually use a generative network that estimates the data distribution of old tasks. Then, the generated pseudo data are added to the arriving new data to jointly train models. For example, BIR \citep{van2020brain} replays the hidden representations generated by the network's own feedback connections. Recently, Cao et al. \citep{cao2024generative} propose to leverage pre-trained generative multi-modal models for class-incremental learning. Different from previous generative replay methods that aim to generate pseudo old data, their approach directly generates label text for a given image using the generative model, and then compares the generated text embeddings with the true label embeddings to make the final classification. However, this kind of methods may fail to generate data points or label text that are fully consistent with the old tasks' distributions on complex datasets, and thus the performance may be degraded.

\subsection{Federated Learning}
In order to solve the problem of the isolation of privacy data, federated learning (FL) \citep{mcmahan2017communication,li2020federated,tan2022fedproto,ma2022layer} has been proposed in recent years and has performed well.
For example, FedAvg \citep{mcmahan2017communication}, which is the most classical method, receives model parameters uploaded by the client on the server side. The parameter values are then simply averaged and returned to each client. FedProx \citep{li2020federated} adds a proximal term on the server side to improve the inconsistency of work performance caused by system heterogeneity between different clients, and help the local models to approximate the global model. pFedLA \citep{ma2022layer} proposes a hierarchical personalized federated learning training framework that exploits inter-user similarities among clients with non-iid data to produce accurate personalized models. 

However, local clients may continuously generate or receive new data. Furthermore, new data may carry new classes which local clients did not know. In such situations, existing federated learning methods could not perform well as most of them assume that data of local clients are fixed \citep{dong2022federated}. To tackle this challenge, we focus on the problem of federated class-incremental learning and design our model to incrementally learn from new data while the catastrophic forgetting of old classes should be well alleviated.

\subsection{Prompt Learning}

Prompt learning is a new transfer learning technique in the field of natural language processing (NLP). The main idea is to learn a function that contains task information to modify the input text, so the language model can get task-relevant information from the input. Prompt tuning \citep{lester2021power,chen2024gap} and prefix tuning \citep{li2021prefix} design prompting functions by applying learnable prompts in a continuous space. Experimental results show that these prompting functions are very effective in transfer learning. In addition, prompts capture more accurate task-specific knowledge with smaller additional parameters than their competitors, such as Adapter \citep{pfeiffer2021adapterfusion,wang2021k} and LoRA \citep{hu2022lora}. 

In recent work, several approaches have been proposed to combine prompt learning with class-incremental learning \citep{smith2023coda,zhang2023complementary,kurniawan2024evolving,li2024steering}. L2P \citep{wang2022learning} combines prompt learning with class-incremental learning to achieve efficient incremental learning without using memory buffers. Building upon L2P, DualPrompt \citep{zhang2023complementary} learns two sets of disjoint prompt spaces, G(eneral)-Prompt and E(xpert)-Prompt, that encode task-invariant and task-specific instructions, respectively. 

While the above works adopt a discrete prompt selection framework based on prompt pools, some recent works have proposed parameterized continuous prompt methods to overcome the limitations of discrete selection. CODA-Prompt \citep{smith2023coda} introduces a decomposed prompt that consists of a weighted sum of learnable prompt components, enabling higher prompting capacity by expanding in a new dimension (the number of components). Furthermore, CODA-Prompt proposes a novel attention-based component-weighting scheme, which allows for end-to-end optimization. EvoPrompt reformulates continual prompting by employing a feed-forward network (FFN) with a multi-layer perceptron (MLP) bottleneck. This approach encodes the prompts in the neural weight space, called prompt memory. EvoPrompt \citep{kurniawan2024evolving} also introduces a dual memory paradigm with a stable reference prompt memory and a flexible working prompt memory. 

Although the effectiveness of prompt learning for class-incremental learning has been shown, it is improper to directly introduce prompt learning into the more complex and practical federated class-incremental learning task. Due to the non-iid problem between clients under the same task, if we directly combine prompt learning and federated learning, the partial classes missing within the task caused by non-iid will further exacerbate the catastrophic forgetting problem of the local and global model.

\begin{figure*}
	\centering
	\includegraphics[width=12cm]{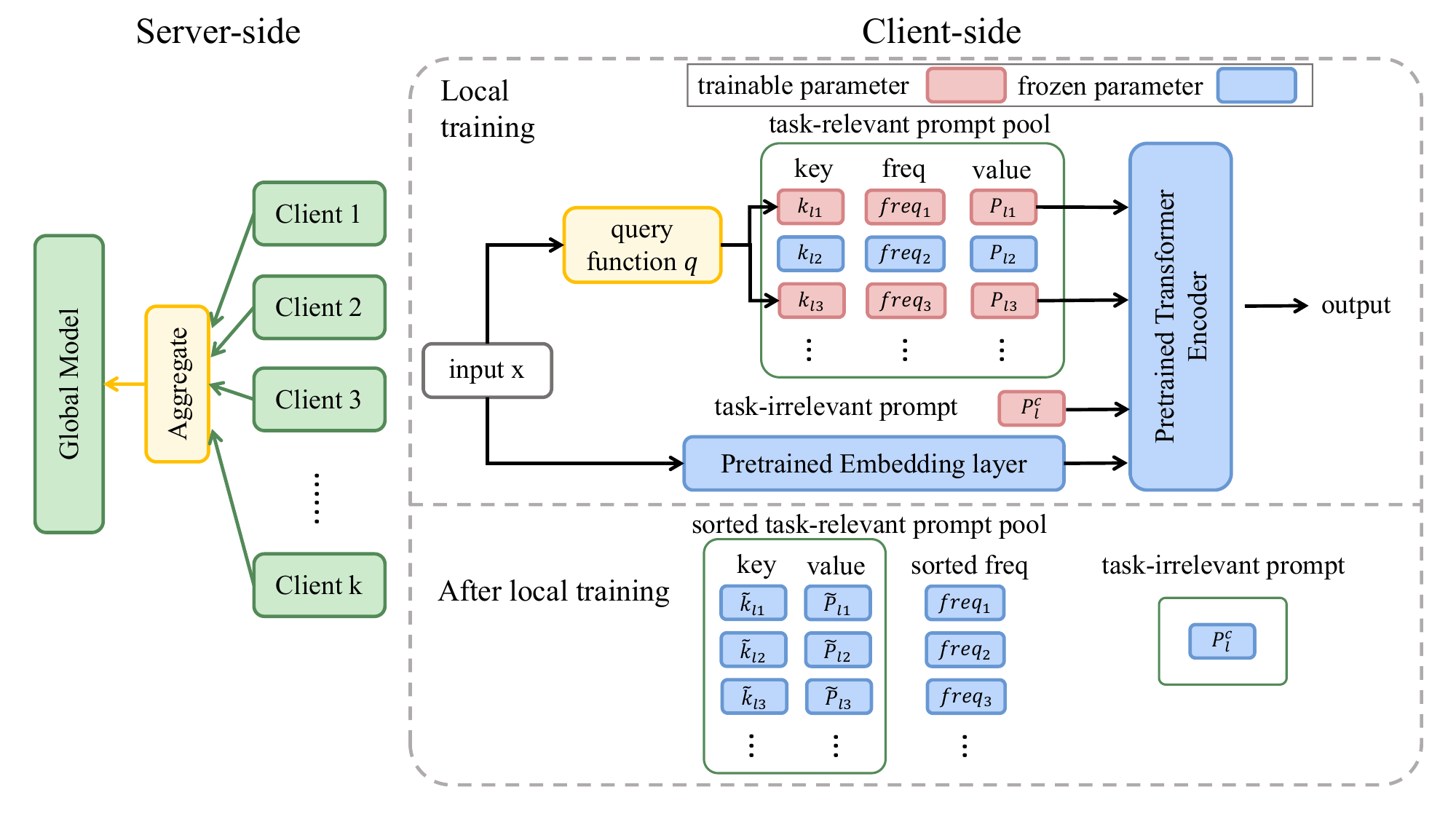}
	\caption{Overview of our method FCILPT. (Left) Illustration of the federated learning workflow, where each client only needs to upload the trainable parameter to the server for aggregation. (Right) Illustration of the detailed pipeline of the local model on the client. During training, the input is passed through the query function q to match prompts in the task-relevant prompt pool. The selected task-relevant prompts, along with task-irrelevant prompts, are fed into the encoder for training. After local training, the prompt pool is sorted in descending order based on the selection frequency of each prompt key, resulting in the locally aligned prompt pool.}
	\label{fig_model}
\end{figure*}

\section{Our Method}
\subsection{Notations and Problem Definition}
In class-incremental learning (CIL), new classes are incrementally fed to the model over time, and the model needs to adapt to the new classes without forgetting the previously learned ones. This is often described as a dynamic arrival task, as the data is non-stationary and arrives continuously over time. In CIL, each task contains data from different classes, and the task identity is usually unknown during testing, making the problem more complex. More formally, CIL can be represented as a sequence of tasks $\mathcal{T}={\{ \mathcal{T}^t \}}_{t=1}^T$ in CIL, where $T$ denotes the number of tasks and the $t$-th task $\mathcal{T}^t={\{ x_i^t, y_i^t \}}_{i=1}^{N^t}$ consists of $N^t$ images $x^t \in \mathcal{X}^t$ and their corresponding labels $y^t \in \mathcal{Y}^t$. Our purpose is to train a model $f(\theta) : \mathcal{X} \rightarrow \mathcal{Y}$ parameterized by $\theta$, such that it can predict the label $y=f(\theta; x) \in \mathcal{Y}$ given a test image $x$ from arbitrary tasks.

We address the problem of extending class-incremental learning to the Federated Learning setting. We consider a scenario where there are $Z$ local clients denoted as $\{{ S_l }\}_{l=1}^Z$ and a global server represented as $S_G$. At each task, we perform $R$ rounds of the aggregation process, where $H$ clients are randomly selected to participate in the aggregation. Specifically, at task $t$, each local client $S_l$ is responsible for training a local model $f_l(\theta_l^t)$ using its local data for the new classes, i.e., $\mathcal{T}_l^t={\{ x_{li}^t, y_{li}^t \}}_{i=1}^{N_l^t} \subset $$\mathcal{T}^t$. The local model parameters are denoted as $\theta_l^t$ and they are updated using the client's data. After training, the local client uploads the updated model parameters to the server $S_G$ for aggregation with other clients' parameters. The server aggregates the uploaded parameters to obtain the global model $f_g(\theta^t)$ for the current round. This process is repeated for $R$ rounds, and the global model's parameters are updated after each round of aggregation.

The Federated Class-Incremental Learning (FCIL) setup proposed by \citep{dong2022federated} involves dividing the local clients $\{{ S_l }\}_{l=1}^Z$ into three categories, namely $S_o$, $S_b$, and $S_n$. Specifically, the clients in $S_o$ do not receive data from the new class in the current task and they have examples only from previous tasks. $S_b$ consists of clients that have data from the current task as well as examples with old data in the memory buffer. Besides, the clients in $S_n$ have not participated in the training of the previous task and they are new clients added under the current task. It is worth noting that the clients in $S_n$ are dynamically changed when a new incremental task arrives. 

However, different from GLFC \citep{dong2022federated}, of which clients use a combination of data from the new task and examples from previous tasks stored in the memory buffer during local training , we propose a rehearsal-free FCIL method that does not rely on memory buffers called Federated Class Incremental Learning with PrompTing (FCILPT). Our approach encodes task-relevant (Sec. \ref{Task-Relevant Prompt}) and task-irrelevant knowledge (Sec. \ref{Task-Irrelevant Prompt}) into prompts, preserving the old and new knowledge of local clients to mitigate catastrophic forgetting. We extend these prompts with learnable keys and utilize an instance-based prompt query mechanism (Sec. \ref{Instance-Wise Prompt Query}) to accurately select suitable prompts for instances, even without prior knowledge of the task identity. To address the non-iid problem caused by the lack of classes between different clients under the new task, FCILPT sorts the selected prompts, aligns task information, and integrates the same task knowledge during global aggregation (Sec. \ref{Aggregating Prompt Pools}). Finally, the overview of our method is depicted in Fig. \ref{fig_model}, and we derive the optimization objective and overall algorithm for FCILPT (Sec. \ref{Overall algorithm of FCILPT}).

\subsection{Task-Relevant Prompt}\label{Task-Relevant Prompt}
At task $t$, the client in $S_o$ does not receive new data, while the $l$-th local client $S_l \in \{S_b \cup S_n\}$ receive training data $\mathcal{T}_l^t$ of new classes. Due to the powerful feature extraction capabilities of pre-trained vision transformer (ViT) \citep{dosovitskiyimage}, we adopt it as local models, i.e., $f_l(\theta_l^t) = h_l(\phi_l^t) \circ g_l(\mu_l^t)$. Here, $g_l(\mu_l^t)$ is the input embedding layer parametrized by $\mu_l^t$, and $h_l(\phi_l^t)$ represents the remaining layers of the ViT parametrized by $\phi_l^t$ (i.e., a stack of self-attention layers and the classification head). The pre-trained embedding layer $g_l(\mu_l^t)$ projects the input images $x_l^t$ to the embedded features $x_{le}^t=g_l(\mu_l^t;x_l^t) \in \mathbb{R}^{L \times D}$, where $L$ is the token length, $D$ is the embedding dimension. Since the ViT has been pre-trained on a large number of images and contains rich image knowledge, retraining the entire model on local data may cause the loss of existing knowledge. Therefore, we freeze the backbone of the pre-trained model.

Learning new classes may cause the model to forget previously learned classes or become biased towards the new classes, resulting in decreased overall performance. To address this issue, many existing CIL methods \citep{rebuffi2017icarl,aljundi2019gradient} utilize a memory buffer to store and select a portion of the data for each old task. When a new task arrives, the samples in the buffer are combined with the arriving new data to train the model together, ensuring that the model retains knowledge of both the old and new tasks. Nonetheless, this approach requires efficient memory management and data selection. To avoid relying on a memory buffer, we introduce a set of learnable parameter called the prompt, which encodes knowledge about each task.

Our method designs a set of task-relevant prompt for each task to encode relevant knowledge, containing two types of prompts: task-specific prompts and task-similar prompts. In each task, some knowledge is unique, which we encode into the prompts as follows:
\begin{equation}\label{eqn-1}
	\begin{aligned}
		&P^{sp}_l = \{ P_{l1}, P_{l2}, \cdots, P_{lT} \},
	\end{aligned}
\end{equation}
where $P_{lj} \in \mathbb{R}^{L_p \times D}$ is a unique prompt for $j$-th task with token length $L_p$ and embedding size $D$, $P^{sp}_l$ is the task-specific prompts pool, and $T$ denotes the number of task. By embedding knowledge into each task-independent prompt, we can learn task-specific knowledge without forgetting old knowledge from previous tasks.

While task-specific prompts allow us to learn unique knowledge for each task, tasks in incremental learning are not entirely independent of each other. Images from different tasks may contain some shared knowledge. If we can identify similar tasks and learn their shared knowledge, it will provide additional learnable knowledge. To leverage this, we increase the size of the local prompt pool and set some additional prompts to learn the shared knowledge between similar tasks, as follows,
\begin{equation}\label{eqn-2}
	\begin{aligned}
		&P_l^{si} = \{P_{lT+1}, P_{lT+2}, \cdots, P_{lM} \},
	\end{aligned}
\end{equation}
where $P^{si}_l$ is the task-similar prompts pool and $M$ is the total number of prompts in the pool. Prompts from $T+1$ to $M$ are used to embed shared knowledge between similar tasks of client $l$.

Combining Eq. \ref{eqn-1} and Eq. \ref{eqn-2}, we get the final prompt pool for the client $l$ as follows,
\begin{equation}\label{eqn-3}
	\begin{aligned}
		&P_l = \{P_{l1}, P_{l2}, \cdots, P_{lT}, P_{lT+1}, P_{lT+2}, \cdots, P_{lM} \},
	\end{aligned}
\end{equation}
where $P_l$ is the prompt pool of client $l$, which encodes task-relevant knowledge to avoid relying on a memory buffer without forgetting old knowledge from previous tasks. 

\subsection{Instance-Wise Prompt Query}\label{Instance-Wise Prompt Query}
As discussed in Section \ref{Task-Relevant Prompt}, the task-relevant prompt pool consists of task-specific and task-similar knowledge shared among all tasks. By using a shared prompts pool for all tasks, we can mitigate interference between old and new tasks and mitigate catastrophic forgetting. During training on a new task, the model can retrieve relevant prompts from the pool and use them to fine-tune the existing parameters. However, since we assume that the task identity of the samples is unknown, we cannot match the input samples to specific prompts without additional operation. Therefore, we propose an instance-based prompt query mechanism, which is inspired by \citep{wang2022learning}, allowing us to select relevant prompts from the pool without knowledge of the task identity.

To perform instance-based prompt querying, we introduce a learnable key-value mapping mechanism. Specifically, we assign a one-to-one learnable key $k_{l}$ to each prompt $P_l$, resulting in a modified prompt pool in the form of key-value pairs, which is as follows, 
\begin{equation}\label{eqn-4}
	\begin{aligned}
		(k_l, P_l) = &\{ (k_{l1}, P_{l1}), (k_{l2}, P_{l2}), \cdots, (k_{lT}, P_{lT}), \\
		&(k_{lT+1}, P_{lT+1}) \cdots, (k_{lM}, P_{lM}) \}.
	\end{aligned}
\end{equation}
The key $k_{li}$ serves as the query identity for the $i$-th prompt value of client $l$, while $P_{li}$ is the $i$-th prompt value of client $l$.

To match input images $x_{l}^t$ to the corresponding prompts in the prompt pool, we use the frozen ViT embedding layer to transform the images into query features $q(x_{l}^t)$. We obtain the embedding features $q(x_{l}^t)$ by inputting the samples into the embedding layer of ViT and using the features of the class token as the query features. Specifically, we use $g_l(\mu_l^t;x_{l}^t)$ to extract the embedding features $q(x_{l}^t)$, and then select the first feature vector (the class token) as the query features of the samples.

Next, we use the query features $q(x_{l}^t)$ to retrieve the most relevant prompts in the prompt pool, i.e., top-$N$ prompts, which consist of a unique task-specific prompt and multiple task-similar prompts of the current task, i.e., $\{P_{ls_1}, P_{ls_2}, \cdots, P_{ls_N}\}$. We determine the top-$N$ prompts using an objective function that calculates the cosine distance between the query features and the prompt keys, which is as follows,
\begin{equation}\label{eqn-5}
	\begin{aligned}
		&\mathcal{L}_{QP} = \mathcal{D}_{DT}(q(x_{l}^t), k_{ls_i}), \\
		&K_{x_{l}^t} = \mathop{\arg\min}_{\{s_i\}_{i=1}^N \subseteq [1, M]} \sum_{i=1}^N \mathcal{L}_{QP},
	\end{aligned}
\end{equation}
where $\{s_i\}_{i=1}^N$ is a subset of indices in the range $[1, M]$ that belong to the $l$-th client. The function $\mathcal{D}_{DT}(\cdot)$ calculates the cosine distance between two vectors, and is used to obtain the match scores between the query features $q(x_l^t)$ and the prompt keys $k_{ls_i}$. $K_{x_{l}^t}$ represents the subset of the top $N$ keys selected from the prompt pool for the sample $x_l^t$.

\subsection{Task-Irrelevant Prompt}\label{Task-Irrelevant Prompt}
While task-specific and task-similar knowledge is important for achieving high performance on individual tasks, capturing common knowledge can enhance the model's ability to generalize to new tasks and improve overall performance in CIL scenarios. Moreover, common knowledge may be valuable for identifying patterns and correlations across tasks. To capture potential common knowledge, we propose a new task-irrelevant prompt $P_l^{c} \in \mathbb{R}^{L_c \times D}$ that is separate from the prompt pool and has a different token length $L_c$. The task-irrelevant prompt aggregates knowledge from all arrived tasks, enabling the model to uncover potentially useful common knowledge.

We combine the prompt pool containing task-specific and task-similar knowledge with the task-irrelevant prompt to incorporate task information into the embedded features of the image. Specifically, the embedded features contain the task-specific prompts $P_{ls_1}$, the shared prompts $P_{ls_2}$ to $P_{ls_N}$, the task-irrelevant prompt $P_l^c$, and the original embedded features $x_{le}^t$, which is as follows,
\begin{equation}\label{eqn-6}
	\begin{aligned}
		&x_{lp}^t = [P_{ls_1}; P_{ls_2}; \cdots; P_{ls_N}; P_l^c; x_{le}^t],
	\end{aligned}
\end{equation}
where $\{s_i\}_{i=1}^N$ is a subset of $N$ indices from $[1, M]$ that belong to the $l$-th client, $x_{lp}^t$ is the expanded features of the image. 

To ensure that the expanded features have discriminative power, we train the model using cross-entropy loss. Specifically, We input the embedded image features with task information into the remaining part of the ViT and obtain the classification loss of the $l$-th local client,
\begin{equation}\label{eqn-7}
	\begin{aligned}
		\mathcal{L}_{CE} = \mathcal{D}_{CE}(h_l(\phi_l^t;x_{lp}^t),y_l^t),
	\end{aligned}
\end{equation}
where $\mathcal{D}_{CE}(\cdot)$ is the cross-entropy loss, $h_l(\phi_l^t)$ represents the remaining layers of the ViT on local client parametrized by $\phi_l^t$ in the $t$-th task, $x_{lp}^t$ is the expanded features of the image at client $l$. By optimizing the model with cross-entropy loss, we can ensure that the embedded features contain task-relevant and task-irrelevant knowledge that is for predicting the task labels. 

Combining Eq.\eqref{eqn-5} and Eq.\eqref{eqn-7} to obtain the overall optimization objective for the $l$-th local client at the $t$-th incremental task as follows,
\begin{equation}\label{eqn-8}
	\begin{aligned}
		\mathcal{L} = \mathcal{L}_{QP} + \lambda \mathcal{L}_{CE},
	\end{aligned}
\end{equation}
The trade-off parameter $\lambda$ is introduced in Eq. \eqref{eqn-8} to balance the contribution of the matching loss $\mathcal{L}_{QP}$ and the classification loss $\mathcal{L}_{CE}$ during optimization. Specifically, the matching loss aims to make the selected keys closer to their corresponding query features, while the classification loss encourages accurate predictions.

\subsection{Aggregating Prompt Pools}\label{Aggregating Prompt Pools}

\begin{algorithm}[t]
	\renewcommand{\algorithmicrequire}{\textbf{Input:}}
	\renewcommand{\algorithmicensure}{\textbf{Output:}}
	\caption{FCILPT Algorithm}
	\label{alg1}
	\begin{algorithmic}[1]
		\REQUIRE  Local clients $\{S_l\}_{l=1}^Z$, local data $\{\mathcal{T}_l\}_{l=1}^Z$, total number of prompts in the prompt pool $M$, the learning rate $\eta$
		\ENSURE Global prompt pool $(k_g, P_g)$ and task-common prompt $P_g^c$
		\renewcommand{\algorithmicrequire}{\textbf{Server executes:}}
		\REQUIRE
		\STATE Initialize global pool $(k_g, P_g)$ and task-common prompt $P_g^c$
		\FOR{each federated  round $ r \in \{ 1,\cdots,R\} $}
		\STATE Server selects a subset $S_r$ of clients $\{S_l\}_{l=1}^Z$ at random
		\FOR{each selected client $l \in S_r$ \textbf{in parallel}}
		\STATE $\hat{k}_l, \hat{P}_l, \hat{P}_l^c \leftarrow LocalUpdate (k_g, P_g, P_g^c) $
		\STATE $(\tilde{k}_l, \tilde{P}_l) \leftarrow sort (\hat{k}_l, \hat{P}_l)$ according to the selection \\ frequency $[freq_1,freq_2,\cdots,freq_M]$
		\ENDFOR		
		\STATE Server aggregation $\quad k_g = \frac{1}{\left| S_r \right|} \sum_{l \in S_r} \tilde{k}_l$ \\
		$P_g = \frac{1}{\left| S_r \right|} \sum_{l \in S_r} \tilde{P}_l, \quad P_g^c = \frac{1}{\left| S_r \right|} \sum_{l \in S_r} \hat{P}_l^c$
		\ENDFOR
		\renewcommand{\algorithmicensure}{\textbf{LocalUpdate}}
		\ENSURE $(k_g, P_g, P_g^c)$:
		\STATE Receive $k_g, P_g, P_g^c$ from server.
		\STATE Updating the local prompts with the global prompts, set \\ $k_l=k_g, P_l=P_g, P_l^c=P_g^c$
		\FOR{each local epoch}
		\FOR{$ (x_l^t, y_l^t) \in \mathcal{T}_l^t$}
		\STATE $\hat{k}_l = k_l - \eta \nabla_{k_l} \mathcal{L}, \qquad \hat{P}_l = P_l - \eta \nabla_{P_l} \mathcal{L},$ \\ $\hat{P}_l^c = P_l^c - \eta \nabla_{P_l^cl} \mathcal{L}$
		\ENDFOR
		\ENDFOR
		\STATE \textbf{return} $\hat{k}_l, \hat{P}_l, \hat{P}_l^c$
	\end{algorithmic}
\end{algorithm}

In federated learning, the non-iid nature of the data distribution amongst clients requires the aggregation of client parameters on the server. As we utilize a pre-trained ViT model with fixed parameters, the main parameters that need to be aggregated are the local task-relevant prompts and the local task-irrelevant prompts of the various clients.

For task-irrelevant prompts, since each client has only one unique prompt, we can simply aggregate them by taking the average over all clients, as follows,
\begin{equation}\label{eqn-9}
	\begin{aligned}
		P_g^c = \frac{1}{\left| S_r \right|} \sum_{l \in S_r} {P}_l^c,
	\end{aligned}
\end{equation}
where $S_r$ is the subset of all clients $\{S_l\}_{l=1}^Z$ selected to participate in local training during the $r$-th global round, $\left| S_r \right|$ is the number of the selected clients $S_r$, ${P}_l^c$ is the local task-irrelevant prompt of $l$-th client, and $P_g^c$ represents task-irrelevant prompts after aggregation.

For task-relevant prompts, our method employs the prompt pool and the order of the prompts in the prompt pool has no specific relationship between clients. However, directly aggregating prompt pools from different clients can result in the aggregation of prompts that embed different task information, thus worsening the non-iid problem. To address this issue, we propose a prompt pool alignment method that reorders prompts in the prompt pool based on their relevance to the current task before aggregation. Our method employs a statistical approach to obtain the frequency of each prompt key's selection in the instance-wise prompt query, denoted as $[freq_1,freq_2,\cdots,freq_M]$. Prompt pool will be sorted in descending order according to the selection frequency of each prompt key and obtain the locally aligned prompt pool. Further, it will be reordered from high to low relevance to the current task, ensuring that prompts in the same position from different clients embed information about the same task and thus addressing the non-iid problem arising from the absence of certain classes under the same task. The reordered prompt pool is as follows,
\begin{equation}\label{eqn-10}
	\begin{aligned}
		&\tilde{k}_l = \{ \tilde{k}_{l1}, \tilde{k}_{l2}, \cdots, \tilde{k}_{lM} \}, \\
		&\tilde{P}_l = \{ \tilde{P}_{l1}, \tilde{P}_{l2}, \cdots, \tilde{P}_{lM} \},
	\end{aligned}
\end{equation}
The size of the prompt pool is denoted as $M$, and each prompt key $k_{li}$ is associated with a corresponding prompt value $P_{li}$.  The prompt values are aligned with their corresponding keys through a one-to-one mapping. The aggregation of the task-relevant prompts and corresponding keys on the server is as follows,
\begin{equation}\label{eqn-11}
	\begin{aligned}
		&k_g = \frac{1}{\left| S_r \right|} \sum_{l \in S_r} \tilde{k}_l, \\
		&P_g = \frac{1}{\left| S_r \right|} \sum_{l \in S_r} \tilde{P}_l, \\
	\end{aligned}
\end{equation}
where $S_r$ is the subset of all clients $\{S_l\}_{l=1}^Z$ that is selected to participate in the local training during the $r$-th global round, $\left| S_r \right|$ is the number of the selected clients $S_r$, $P_g$ is the task-relevant prompts after aggregation, and $k_g$ is corresponding keys after aggregation. 

\subsection{Overall algorithm of FCILPT}\label{Overall algorithm of FCILPT}

The federated global round of the proposed FCILPT is illustrated in Algorithm \ref{alg1}. 
1)Firstly, the server randomly selects a subset $S_r$ of clients ${S_l}_{l=1}^Z$ to participate in the next round of local training. Then, the server transmits the task-relevant prompt pool $(k_g, P_g)$ and the task-irrelevant prompt $P_g^c$ that were previously aggregated on the server to the selected local clients.
2) After receiving the aggregated prompts, each client updates its local task-relevant prompt pool and task-irrelevant prompt. The client queries the prompts in the pool that match the current task and adds the prompts $P_{ls_i}$ obtained from querying to the embedding features of the data along with the task-irrelevant prompt $P_l^c$. 
3) The embedding features $x_{lp}$ containing task information are input into the remaining layers of the ViT, and local training is performed using Eq.\eqref{eqn-7} to obtain the updated local prompt pool $(\hat{k}_l, \hat{P}_l)$ and task-irrelevant prompt $\hat{P}_l^c$.
4) The prompt pool is then sorted based on the times of each prompt is selected, resulting in a task-aligned prompt pool $(\tilde{k}_l, \tilde{P}_l)$. 
5) The selected clients upload their local task-relevant prompt pools and task-irrelevant prompts to the server, and the server aggregates the parameters to obtain the global task-relevant prompt pool $(k_g, P_g)$ and task-irrelevant prompt $P_g^c$.

\section{Experiments}
\subsection{Experimental Settings}
\subsubsection{Datasets} 
We conducted experiments on the following three datasets: CIFAR-100 \citep{krizhevsky2009learning}, Mini-ImageNet \citep{deng2009imagenet}, and Tiny-ImageNet \citep{le2015tiny}.
\textbf{CIFAR-100} is a labeled subset of a set of 80 million natural images for object recognition. The dataset is composed of 60,000 color images from 100 classes with 500 images for training and 100 images for testing. All the images are set to the size of 32 $\times$ 32. \textbf{Mini-ImageNet}  is a subset of ImageNet, and it includes 100 classes. We split each class into 500 training and 100 test samples with the size 224 $\times$ 224. \textbf{Tiny-ImageNet} is also a subset of ImageNet according to the semantic hierarchy. The dataset has 200 classes and each class has 500 training images, 50 validation images, and 50 testing images. All the images are resized to 64 $\times$ 64. 

\subsubsection{Competing Methods and Implementation Details}
To validate the effectiveness of our FCILPT, we compared it with several competing methods, i.e., iCaRL \citep{rebuffi2017icarl}, BiC \citep{wu2019large}, LUCIR \citep{hou2019learning}, PODNet \citep{douillard2020podnet}, DDE \citep{hu2021distilling}, GeoDL \citep{simon2021learning}, SS-IL \citep{ahn2021ss}, AFC \citep{kang2022class}, GLFC \citep{dong2022federated}, and LGA \citep{dong2024no}. 

Our method is based on pre-trained ViT-B/16 \citep{dosovitskiyimage, zhang2022nested}, which has become a widely-used backbone in literature. For baseline methods, their original backbones are ResNet18 \citep{he2016deep}. In this paper, we first compared our method with original baselines with ResNet18 and the results of baselines are directly borrowed from \citep{dong2022federated} as we used the same setting. Then, for fair comparisons, we also replaced baselines' backbones with ViT-B/16 and obtained the results. The source codes of most baselines are kindly provided by the authors. For those methods without publicly available code, we carefully implemented them by ourselves.

We trained our model using Adam \citep{kingma2014method} with a batch size of 16, and a constant learning rate of 0.03. The exemplar memory of each client is set as 2,000 for all streaming tasks. Input images are resized to 224 $\times$ 224 and normalized to the range of [0, 1] to match the pretraining setting. We set $M = 10, N = 5, L_p = 5, L_c = 5$ for CIFAR-100. For Mini-ImageNet, we use $M = 19, N = 4, L_p = 6, L_c = 5$. And for Tiny-ImageNet, we set $M = 24, N = 8, L_p=5, L_c = 1$.

\begin{table*}[t]\footnotesize
	\centering
	\caption{The experimental results on CIFAR-100 with 10 incremental tasks. Baselines use ResNet-18 as backbones.}
	\label{tab:1}{
            \resizebox{\textwidth}{!}{
            \begin{tabular}{ccccccccccccc}
			\toprule
			Methods & 10 & 20 & 30 & 40 & 50 & 60 & 70 & 80 & 90 & 100 & average & performance gains \\
			\midrule
			iCaRL \citep{rebuffi2017icarl}+FL & 89.0 & 55.0 & 57.0 & 52.3 & 50.3 & 49.3 & 46.3 & 41.7 & 40.3 & 36.7 & 51.8 & $\uparrow$ 38.6  \\
			BiC \citep{wu2019large}+FL & 88.7 & 63.3 & 61.3 & 56.7 & 53.0 & 51.7 & 48.0 & 44.0 & 42.7 & 40.7 & 55.0 & $\uparrow$ 35.4 \\
			PODNet \citep{douillard2020podnet}+FL & 89.0 & 71.3 & 69.0 & 63.3 & 59.0 & 55.3 & 50.7 & 48.7 & 45.3 & 45.0 & 59.7 & $\uparrow$ 30.7 \\
			DDE \citep{hu2021distilling}+iCaRL+FL & 88.0 & 70.0 & 67.3 & 62.0 & 57.3 & 54.7 & 50.3 & 48.3 & 45.7 & 44.3 & 58.8 & $\uparrow$ 31.6 \\
			GeoDL \citep{simon2021learning}+iCaRL+FL & 87.0 & 76.0 & 70.3 & 64.3 & 60.7 & 57.3 & 54.7 & 50.3 & 48.3 & 46.3 & 61.5 & $\uparrow$ 28.9 \\
			SS-IL \citep{ahn2021ss}+FL & 88.3 & 66.3 & 54.0 & 54.0 & 44.7 & 54.7 & 50.0 & 47.7 & 45.3 & 44.0 & 54.9 & $\uparrow$ 35.5 \\
   			DyTox \citep{Douillard2022DyTox}+FL & 86.2 & 76.9 & 73.3 & 69.5 & 62.1 & 62.7 & 58.1 & 57.2 & 55.4 & 52.1 & 65.4 & $\uparrow$ 25.0 \\
			AFC \citep{kang2022class}+FL & 85.6 & 73.0 & 65.1 & 62.4 & 54.0 & 53.1 & 51.9 & 47.0 & 46.1 & 43.6 & 58.2 & $\uparrow$ 32.2 \\
			GLFC \citep{dong2022federated}& 90.0 & 82.3 & 77.0 & 72.3 & 65.0 & 66.3 & 59.7 & 56.3 & 50.3 & 50.0 & 66.9 & $\uparrow$ 23.5 \\
                LGA \citep{dong2024no}& 89.6 & 83.2 & 79.3 & 76.1 & 72.9 & 71.7 & 68.4 & 65.7 & 64.7 & 62.9 & 73.5 & $\uparrow$ 16.9 \\
			FCILPT(ours) & \textbf{99.1} & \textbf{96.8} & \textbf{93.3} & \textbf{91.6} & \textbf{89.9} & \textbf{88.9} & \textbf{87.0} & \textbf{85.9} & \textbf{85.9} & \textbf{85.9} & \textbf{90.4} & -- \\
			\bottomrule
	\end{tabular}	
            }
    }
\end{table*}

\begin{table*}[t]\footnotesize
	\centering
	\caption{The experimental results on Mini-ImageNet with 10 incremental tasks. Baselines use ResNet-18 as backbones.}
	\label{tab:2}{
            \resizebox{\textwidth}{!}{
            \begin{tabular}{ccccccccccccc}
			\toprule
			Methods & 10 & 20 & 30 & 40 & 50 & 60 & 70 & 80 & 90 & 100 & average & performance gains \\
			\midrule
			iCaRL \citep{rebuffi2017icarl}+FL & 74.0 & 62.3 & 56.3 & 47.7 & 46.0 & 40.3 & 37.7 & 34.3 & 33.3 & 32.7 & 46.5 & $\uparrow$ 47.4  \\
			BiC \citep{wu2019large}+FL & 74.3 & 63.0 & 57.7 & 51.3 & 48.3 & 46.0 & 42.7 & 37.7 & 35.3 & 34.0 & 49.0 & $\uparrow$ 44.9 \\
			PODNet \citep{douillard2020podnet}+FL & 74.3 & 64.0 & 59.0 & 56.7 & 52.7 & 50.3 & 47.0 & 43.3 & 40.0 & 38.3 & 52.6 & $\uparrow$ 41.3 \\
			DDE \citep{hu2021distilling}+iCaRL+FL & 76.0 & 57.7 & 58.0 & 56.3 & 53.3 & 50.7 & 47.3 & 44.0 & 40.7 & 39.0 & 52.3 & $\uparrow$ 41.6 \\
			GeoDL \citep{simon2021learning}+iCaRL+FL & 74.0 & 63.3 & 54.7 & 53.3 & 50.7 & 46.7 & 41.3 & 39.7 & 38.3 & 37.0 & 50.0 & $\uparrow$ 43.9 \\
			SS-IL \citep{ahn2021ss}+FL & 69.7 & 60.0 & 50.3 & 45.7 & 41.7 & 44.3 & 39.0 & 38.3 & 38.0 & 37.3 & 46.4 & $\uparrow$ 47.5 \\
      		DyTox \citep{Douillard2022DyTox}+FL & 76.3 & 68.3 & 64.8 & 58.6 & 45.4 & 41.3 & 39.7 & 37.1 & 36.2 & 35.3 & 50.3 & $\uparrow$ 43.6 \\
			AFC \citep{kang2022class}+FL & 82.5 & 74.1 & 66.8 & 60.0 & 48.0 & 44.3 & 42.5 & 40.9 & 39.0 & 36.1 & 53.4 & $\uparrow$ 40.5 \\
			GLFC \citep{dong2022federated}& 73.0 & 69.3 & 68.0 & 61.0 & 58.3 & 54.0 & 51.3 & 48.0 & 44.3 & 42.7 & 57.0 & $\uparrow$ 36.9 \\
                LGA \citep{dong2024no}& 83.0 & 74.2 & 72.3 & 72.2 & 68.1 & 65.8 & 64.0 & 59.6 & 58.4 & 57.5 & 67.5 & $\uparrow$ 26.4 \\
			FCILPT(ours) & \textbf{98.3} & \textbf{96.9} & \textbf{95.3} & \textbf{95.0} & \textbf{93.8} & \textbf{93.1} & \textbf{93.2} & \textbf{92.0} & \textbf{90.8} & \textbf{90.4} & \textbf{93.9} & -- \\
			\bottomrule
	\end{tabular}	
            }
		
        }
\end{table*}

\begin{table*}[t] \footnotesize
	\centering
	\caption{The experimental results on Tiny-ImageNet with 10 incremental tasks. Baselines use ResNet-18 as backbones.}
	\label{tab:3}{
            \resizebox{\textwidth}{!}{
            \begin{tabular}{ccccccccccccc}
			\toprule
			Methods & 20 & 40 & 60 & 80 & 100 & 120 & 140 & 160 & 180 & 200 & average &performance gains\\
			\midrule
			iCaRL \citep{rebuffi2017icarl}+FL & 63.0 & 53.0 & 48.0 & 41.7 & 38.0 & 36.0 & 33.3 & 30.7 & 29.7 & 28.0 & 40.1 & $\uparrow$ 46.3  \\
			BiC \citep{wu2019large}+FL & 65.3 & 52.7 & 49.3 & 46.0 & 40.3 & 38.3 & 35.7 & 33.0 & 32.7 & 29.0 & 42.1 & $\uparrow$ 44.3 \\
			PODNet \citep{douillard2020podnet}+FL & 66.7 & 53.3 & 50.0 & 47.3 & 43.7 & 42.7 & 40.0 & 37.3 & 33.7 & 31.3 & 44.6 & $\uparrow$ 41.8 \\
			DDE \citep{hu2021distilling}+iCaRL+FL & 69.0 & 52.0 & 50.7 & 47.0 & 43.3 & 42.0 & 39.3 & 37.0 & 33.0 & 31.3 & 44.5 & $\uparrow$ 41.9 \\
			GeoDL \citep{simon2021learning}+iCaRL+FL & 66.3 & 54.3 & 52.0 & 48.7 & 45.0 & 42.0 & 39.3 & 36.0 & 32.7 & 30.0 & 44.6 & $\uparrow$ 41.8 \\
			SS-IL \citep{ahn2021ss}+FL & 62.0 & 48.7 & 40.0 & 38.0 & 37.0 & 35.0 & 32.3 & 30.3 & 28.7 & 27.0 & 37.9 & $\uparrow$ 48.5 \\
         	DyTox \citep{Douillard2022DyTox}+FL & 73.2 & 66.6 & 48.0 & 47.1 & 41.6 & 40.8 & 37.4 & 36.2 & 32.8 & 30.6 & 45.4 & $\uparrow$ 41.0 \\
			AFC \citep{kang2022class}+FL & 73.7 & 59.1 & 50.8 & 43.1 & 37.0 & 35.2 & 32.6 & 32.0 & 28.9 & 27.1 & 42.0 & $\uparrow$ 44.4 \\
			GLFC \citep{dong2022federated}& 66.0 & 58.3 & 55.3 & 51.0 & 47.7 & 45.3 & 43.0 & 40.0 & 37.3 & 35.0 & 47.9 & $\uparrow$ 38.5 \\
                LGA \citep{dong2024no}& 70.3 & 64.0 & 60.3 & 58.0 & 55.8 & 53.1 & 47.9 & 45.3 & 39.8 & 37.3 & 53.2 & $\uparrow$ 33.2 \\
			FCILPT(ours) & \textbf{89.7} & \textbf{89.0} & \textbf{87.2} & \textbf{88.4} & \textbf{87.2} & \textbf{86.7} & \textbf{85.4} & \textbf{84.2} & \textbf{83.7} & \textbf{82.5} & \textbf{86.4} & -- \\
			\bottomrule
	\end{tabular}	
            }
        }
\end{table*}

\begin{table*}[t] \footnotesize
	\centering
	\caption{The experimental results on CIFAR-100 with 5 incremental tasks. Baselines use ViT-B/16 as backbones. }
	\label{tab:4}{
		\begin{tabular}{cccccccc}
			\toprule
			Methods & 20 & 40 & 60 & 80 & 100 & average & performance gains \\
			\midrule
			iCaRL \citep{rebuffi2017icarl}+FL  & 91.8 & 83.7 & 77.6 & 72.7 & 66.0 & 78.3 & $\uparrow$ 12.5  \\
			BiC \citep{wu2019large}+FL & 92.3 & 78.8 & 74.3 & 72.7 & 64.2 & 76.4 & $\uparrow$ 14.4 \\
			LUCIR \citep{hou2019learning}+FL & 91.7 & 76.9 & 73.4 & 74.1 & 66.9 & 76.6 & $\uparrow$ 14.2 \\
			GLFC \citep{dong2022federated} & 88.2 & 77.9 & 72.6 & 67.7 & 58.4 & 73.0 & $\uparrow$ 17.8 \\
			FCILPT (ours) & \textbf{96.3} & \textbf{92.8} & \textbf{89.8} & \textbf{87.9} & \textbf{87.2} & \textbf{90.8} & -- \\
			\bottomrule
	\end{tabular}	}
\end{table*}

\begin{table*}[t]\footnotesize
	\centering
	\caption{The experimental results on CIFAR-100 with 10 incremental tasks. Baselines use ViT-B/16 as backbones.}
	\label{tab:5}{
        \resizebox{\textwidth}{!}{
        \begin{tabular}{ccccccccccccc}
			\toprule
			Methods & 10 & 20 & 30 & 40 & 50 & 60 & 70 & 80 & 90 & 100 & average & performance gains \\
			\midrule
			iCaRL \citep{rebuffi2017icarl}+FL  & 95.6 & 90.8 & 85.6 & 83.2 & 79.4 & 77.0 & 75.5 & 73.7 & 72.6 & 70.1 & 80.3 & $\uparrow$ 10.1  \\
			BiC \citep{wu2019large}+FL & 95.7 & 87.5 & 84.8 & 80.8 & 76.9 & 77.4 & 77.3 & 75.2 & 73.0 & 70.6 & 79.9 & $\uparrow$ 10.5 \\
			LUCIR \citep{hou2019learning}+FL & 95.0 & 87.2 & 81.6 & 80.8 & 78.0 & 76.3 & 77.3 & 72.3 & 71.6 & 71.5 & 79.1 & $\uparrow$ 11.3 \\
			GLFC \citep{dong2022federated} & 93.7 & 87.8 & 75.9 & 69.3 & 62.7 & 61.2 & 56.7 & 55.2 & 52.3 & 44.5 & 65.9 & $\uparrow$ 24.5 \\
			FCILPT (ours) & \textbf{99.1} & \textbf{96.8} & \textbf{93.3} & \textbf{91.6} & \textbf{89.9} & \textbf{88.9} & \textbf{87.0} & \textbf{85.9} & \textbf{85.9} & \textbf{85.9} & \textbf{90.4} & -- \\
			\bottomrule
	\end{tabular}	
        }
    }
\end{table*}

\subsubsection{Settings of Federated Class-Incremental Learning.}
This paper follows the experimental settings in \citep{dong2022federated}. 

For the federated learning setting, there are 30 local clients in the first task. For each global round, we randomly selected 10 clients to conduct 5-epoch local training. After the local training, these clients will share their updated models to participate in the global aggregation of this round. (1) When the number of streaming tasks is $T = 10$, on CIFAR-100 and Mini-ImageNet, each task includes 10 new classes for 10 global rounds, and each task transition will introduce 10 additional new clients. On Tiny-ImageNet, each task includes 20 new classes for the same 10 global rounds, and each task transition also includes 10 new clients. (2) For the case of streaming tasks $T = 5$, each task has 20 classes with 20 global rounds of training on CIFAR-100 and Mini-ImageNet, and the number of classes will be 40 on Tiny-ImageNet. Note that the number of newly introduced clients is 20 now at each task transition. (3) We also conducted experiments of $T = 20$. CIFAR-100 and Mini-ImageNet contain 5 classes for each task, while Tiny-ImageNet has 10 classes per task. For all three datasets, each task covers 10 global rounds and there will be 5 new clients joining in the framework at each task transition.

For building the non-iid setting, every client can only own 60\% classes of the label space in the current task, and these classes are randomly selected. During the task transition global round, we assumed that 90\% existing clients are from $S_b$, while the resting 10\% clients are from $S_o$.

\subsubsection{Evaluation Metrics}
Following \citep{dong2022federated}, we also used the Average Accuracy ($A$) to evaluate the overall performance. The average accuracy measures the model's classification performance at every phase after learning a new task. Avg.(\%) is the averaged performance of all phases. The $t$-th accuracy is defined as  $A_t = \frac{1}{t} \sum_{j=1}^t a_{t,j}$, where $a_{t,j} (j \leq t)$ is the accuracy of task $j$ in phase $t$.

\subsection{Experimental Results}
\subsubsection{Using Original Backbones for Baselines}
\begin{figure}[h]
	\centering
	\includegraphics[width=0.8\linewidth]{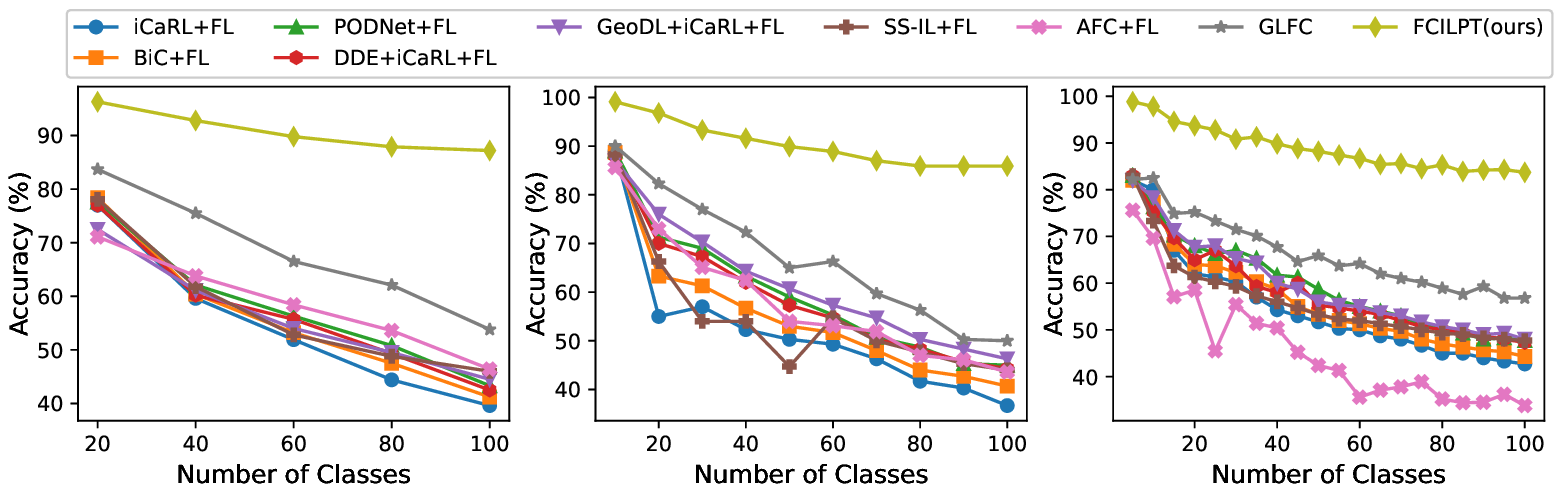}
	\caption{Qualitative analysis of different incremental tasks on CIFAR-100 when T = 5 (left), T = 10 (middle), and T = 20 (right). Baselines use ResNet18 as backbones.}
	\label{fig_2}
\end{figure}

\begin{figure}[h]
	\centering
	\includegraphics[width=0.8\linewidth]{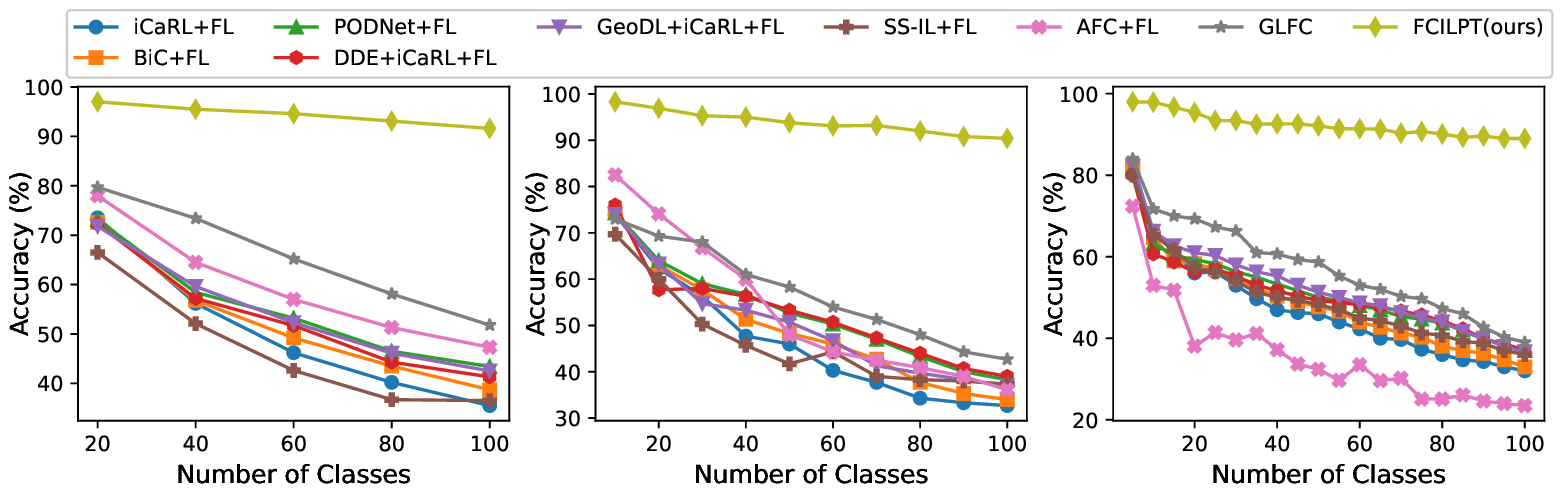}
	\caption{Qualitative analysis of different incremental tasks on Mini-Imagenet when T = 5 (left), T = 10 (middle), and T = 20 (right). Baselines use ResNet18 as backbones.}
	\label{fig_3}
\end{figure}

\begin{figure}[h]
	\centering
	\includegraphics[width=0.8\linewidth]{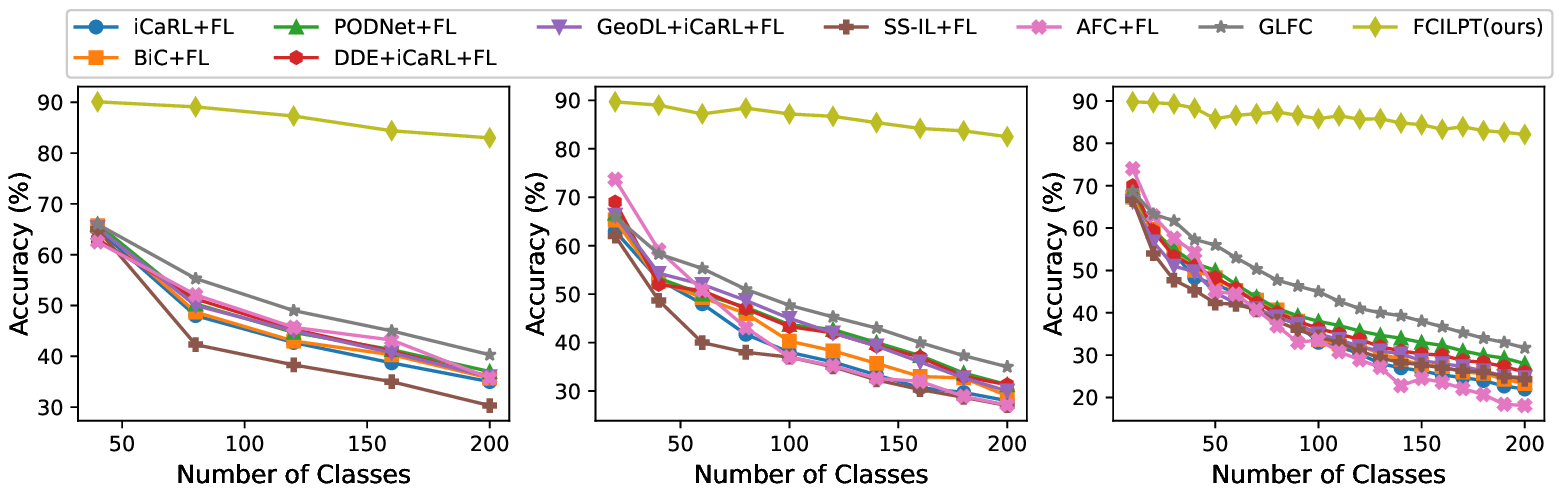}
	\caption{Qualitative analysis of different incremental tasks on Tiny-Imagenet when T = 5 (left), T = 10 (middle), and T = 20 (right). Baselines use ResNet18 as backbones.}
	\label{fig_4}
\end{figure}

In this section, we kept baselines unchanged and used their original backbones ResNet-18 for comparison. The results on CIFAR-100, Mini-ImageNet, and Tiny-ImageNet are reported in Table \ref{tab:1}, Table \ref{tab:2}, and Table \ref{tab:3}, respectively. As shown in Figs. \ref{fig_2}, \ref{fig_3}, \ref{fig_4}, we present comparison between our proposed method with other baseline methods under three incremental tasks on three benchmark datasets. From these results, we could find that traditional incremental methods suffer significant performance degradation in the federated learning setting. Especially, traditional methods like iCaRL and BiC still achieve decent performance on the first task while their performance drop a lot when the number of learned tasks increases.This phenomenon indicates that simply combining incremental learning methods with federated learning leads to catastrophic forgetting and non-iid data distribution among different clients. GLFC is specially designed for federated class-incremental learning problem and is the strongest baseline. However, its performance is still far worse than our method FCILPT. 

The proposed FCILPT could significantly outperform all baselines, showing its effectiveness. 

\begin{figure}
	\centering
	\includegraphics[width=0.8\linewidth]{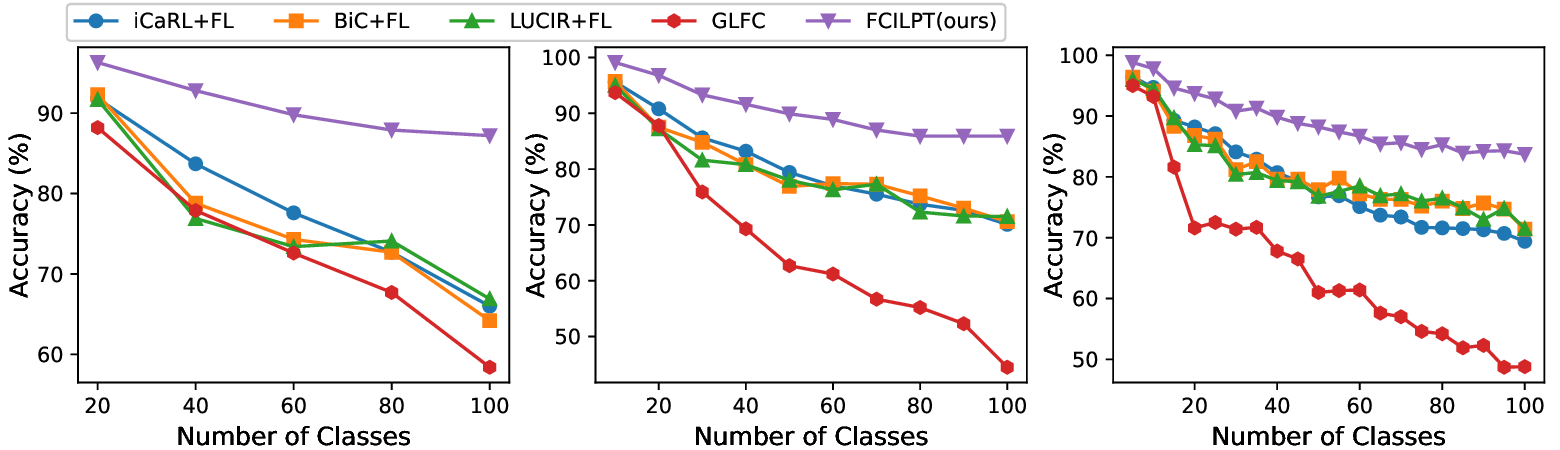}
	\caption{Qualitative analysis of different incremental tasks on CIFAR-100 when T = 5 (left), T = 10 (middle), and T = 20 (right).  Baselines use ViT-B/16 as backbones.}
	\label{fig_5}
\end{figure}

\begin{figure}
	\centering
	\includegraphics[width=0.8\linewidth]{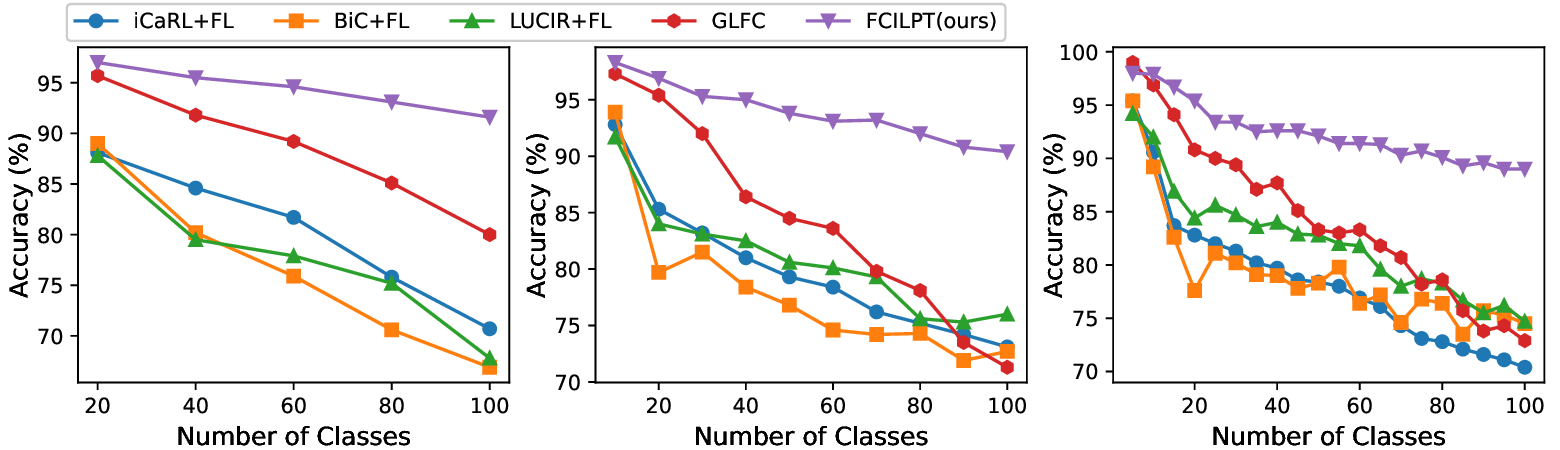}
	\caption{Qualitative analysis of different incremental tasks on Mini-Imagenet when T = 5 (left), T = 10 (middle), and T = 20 (right). Baselines use ViT-B/16 as backbones.}
	\label{fig_6}
\end{figure}

\begin{figure}
	\centering
	\includegraphics[width=0.8\linewidth]{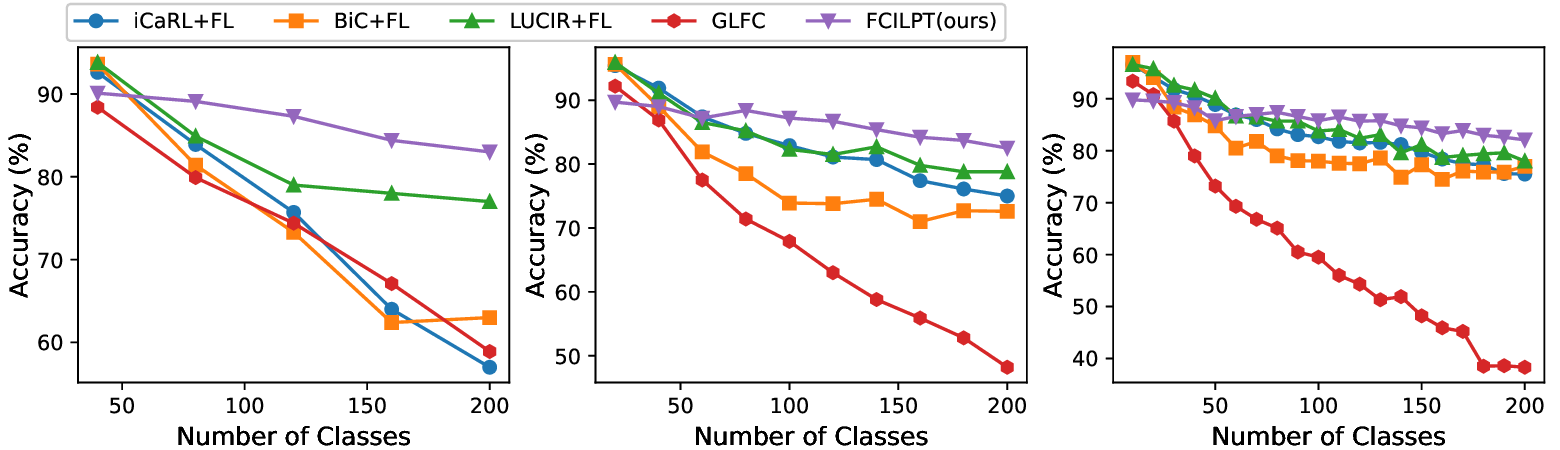}
	\caption{Qualitative analysis of different incremental tasks on Tiny-Imagenet when T = 5 (left), T = 10 (middle), and T = 20 (right). Baselines use ViT-B/16 as backbones.}
	\label{fig_7}
\end{figure}

\subsubsection{Using ViT-B/16 Backbones for Baselines} \label{Further Fair Experiments}
The above results of baseliens are based on ResNet-18 while our method uses ViT-B/16. To ensure fairness and thoroughly validate the effectiveness of our method FCILPT, we replaced the backbones of baselines with ViT-B/16. Experimental results on three datasets are illustrated in Figs. \ref{fig_5}, \ref{fig_6}, \ref{fig_7}. As shown in Tab. \ref{tab:4}, \ref{tab:5}, \ref{tab:6}, the detailed average accuracy in each period of three incremental tasks on CIFAR-100 are presented. From these results, we can find that most baselines could perform better benefiting from using  ViT-B/16 as backbones. Our prposed method could still achieve the best performance in most cases as FCILPT embeds different knowledge into the prompts while retaining knowledge from both old and new tasks, effectively addressing catastrophic forgetting problem.

\begin{table*}[t]\footnotesize
	\centering
	\caption{The experimental results on CIFAR-100 with 20 incremental tasks. Baselines use ViT-B/16 as backbones.}
	\label{tab:6}
	\resizebox{\linewidth}{!}{
		\begin{tabular}{ccccccccccccccccccccccc}
			\toprule
			Methods & 5 & 10 & 15 & 20 & 25 & 30 & 35 & 40 & 45 & 50 & 55 & 60 & 65 & 70 & 75 & 80 & 85 & 90 & 95 & 100 & average & performance gains \\
			\midrule
			iCaRL \citep{rebuffi2017icarl}+FL  & 96.4 & 94.7 & 89.3 & 88.2 & 87.1 & 84.1 & 82.9 & 80.7 & 79.3 & 76.7 & 76.9 & 75.1 & 73.7 & 73.4 & 71.7 & 71.6 & 71.5 & 71.3 & 70.7 & 69.4 & 79.2 & $\uparrow$ 9.7  \\
			BiC \citep{wu2019large}+FL & 96.4 & 94.1 & 88.3 & 86.8 & 86.2 & 81.2 & 82.5 & 79.6 & 79.6 & 77.9 & 79.8 & 77.2 & 76.3 & 76.3 & 75.2 & 76.0 & 74.8 & 75.7 & 74.7 & 71.4 & 80.5 & $\uparrow$ 8.4 \\
			LUCIR \citep{hou2019learning}+FL & 96.0 & 94.5 & 89.7 & 85.3 & 85.1 & 80.4 & 80.7 & 79.4 & 79.2 & 76.8 & 77.6 & 78.5 & 76.9 & 77.2 & 76.0 & 76.5 & 74.9 & 73.0 & 74.8 & 71.5 & 80.2 & $\uparrow$ 8.7 \\
			GLFC \citep{dong2022federated} & 95.0 & 93.2 & 81.6 & 71.6 & 72.5 & 71.4 & 71.7 & 67.8 & 66.5 & 61.0 & 61.3 & 61.4 & 57.6 & 57.0 & 54.6 & 54.2 & 51.9 & 52.3 & 48.7 & 48.8 & 65.0 &   $\uparrow$   23.9 \\
			FCILPT (ours) & \textbf{98.8} & \textbf{97.8} & \textbf{94.6} & \textbf{93.7} & \textbf{92.8} & \textbf{90.8} & \textbf{91.3} & \textbf{89.8} & \textbf{88.8} & \textbf{88.2} & \textbf{87.4} & \textbf{86.7} & \textbf{85.4} & \textbf{85.6} & \textbf{84.5} & \textbf{85.3} & \textbf{83.9} & \textbf{84.2} & \textbf{84.3} & \textbf{83.7} & \textbf{88.9} & -- \\
			\bottomrule
	\end{tabular}	}
\end{table*}

To conclude, our method is effective to handle the the practical and challenging federated class-incremental learning (FCIL) problem.

\subsection{Ablation Study}

\begin{table*}[t] \footnotesize
	\centering
	\caption{Ablation studies on CIFAR-100 with 10 incremental tasks}
	\label{tab:13}{
		\begin{tabular}{cccccccccccccc}
			\toprule
			sorted & task-relevant & task-irrelevant & 10 & 20 & 30 & 40 & 50 & 60 & 70 & 80 & 90 & 100 & average \\
			\midrule
			& & & 95.2 & 88.9 & 85.1 & 81.3 & 78.9 & 76.6 & 75.6 & 73.8 & 73.0 & 71.2 & 80.0 \\
			& \checkmark & & 98.9 & 96.4 & 93.2 & 91.1 & 89.3 & 88.3 & 86.6 & 85.3 & 85.2 & 84.9 & 89.9 \\
			& & \checkmark & 97.8 & 93.9 & 92.1 & 90.5 & 88.6 & 87.2 & 86.3 & 85.5 & 85.0 & 84.3 & 89.1 \\
			& \checkmark & \checkmark & \textbf{99.1} & 96.7 & 93.0 & 91.3 & 89.8 & 88.5 & 86.6 & 85.7 & 85.5 & 85.3 & 90.1 \\
			\checkmark & \checkmark & & 99.0 & 96.4 & 92.7 & 91.2 & 89.8 & 88.7 & 86.9 & 85.8 & \textbf{85.9} & 85.6 & 90.2 \\
			\checkmark & \checkmark & \checkmark & \textbf{99.1} & \textbf{96.8} & \textbf{93.3} & \textbf{91.6} & \textbf{89.9} & \textbf{88.9} & \textbf{87.0} & \textbf{85.9} & \textbf{85.9} & \textbf{85.9} & \textbf{90.4} \\
			\bottomrule
	\end{tabular}	}
\end{table*}

\begin{figure}
	\centering 
	\includegraphics[width=0.8\linewidth]{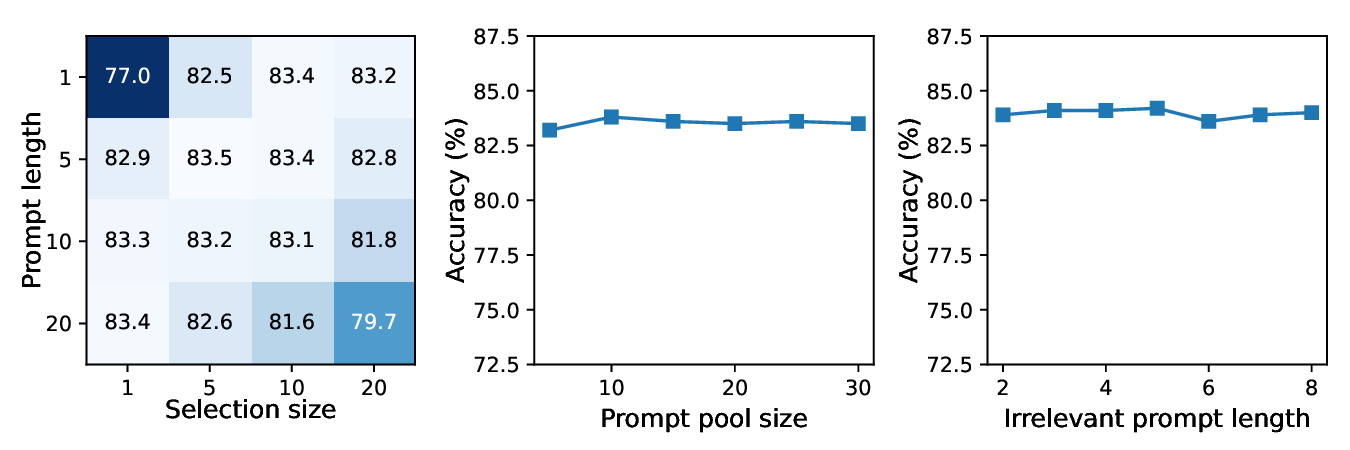}
	\caption{Results on CIFAR-100. Left: Accuracy (\%) w.r.t prompt length $L_p$ and prompt selection size $N$, given $M$ = 20. Middle: Accuracy (\%) w.r.t. prompt pool size $M$, given $L_p$ = 5, $N$ = 5. Right: Accuracy (\%) w.r.t. irrelevant prompt length $L_c$, given $L_p$ = 5, $N$ = 5, $M$ = 10.}
	\label{fig_8}
\end{figure}

To further verify the effectiveness of each proposed component separately, we conducted ablation experiments on CIFAR-100. As shown in Table \ref{tab:13}, after removing the sorting module of the prompt pool, the performance decreases by 0.3\% and 0.2\%, respectively. This phenomenon indicates that sorting the prompt pool before aggregating it from different clients aligns the information related to the same task and alleviates the non-iid distributed problem caused by missing classes. When task-relevant prompts were removed, a significant drop in the model's performance is observed. Task-relevant prompts contain task-specific knowledge and the knowledge shared among similar tasks. Removing these prompts results in the model losing clear and explicit preservation of old knowledge, leading to catastrophic forgetting. Task-irrelevant prompts facilitate learning common knowledge shared across all tasks, further improving the performance of federated incremental learning.

\subsection{Hyperparameters Tuning}
Our method utilizes the prompt learning technique which involves several hyperparameters (including the length of single prompt $L_p$, the size of the prompt pool $M$, the number of prompts selected each time $N$, and the length of irrelevant prompt $L_c$). Taking the CIFAR-100 dataset as an example, we conducted experiments on these hyperparameters in a central learning scenario and presented the results in Fig. \ref{fig_8}. From the left sub-picture, we could find that the highest accuracy is achieved when both $L_p$ and $N$ are set to 5. In the middle sub-picture, different values of parameter M have little effect on the results. From the right sub-picture, it could be found that accuracy is robust to the values of $L_c$.

\section{Conclusion}
In this paper, we propose a rehearsal-free FCIL method that does not rely on memory buffers called Federated Class Incremental Learning with PrompTing (FCILPT). FCILPT encodes task-relevant and task-irrelevant knowledge into prompts, preserving the old and new knowledge of local clients to mitigate catastrophic forgetting. We extend these prompts with learnable keys and utilize an instance-based prompt query mechanism to accurately select suitable prompts for instances, even without prior knowledge of the task identity. To address the non-iid problem caused by the lack of classes between different clients under the new task, FCILPT sorts the selected prompts, aligns task information, and integrates the same task knowledge during global aggregation. Experiments on CIFAR-100, ImageNet-Subset, and TinyImageNet show that FCILPT achieves significant accuracy improvement compared to the state-of-the-art FCIL methods.

\section*{CRediT authorship contribution statement}
\textbf{Xin Luo:} Writing - review \& editing, Writing - original draft, Visualization, Validation, Supervision, Resources, Project administration, Methodology, Investigation, Formal analysis, Data curation, Conceptualization. \textbf{Fang-Yi Liang:} Writing - review \& editing, Writing - original draft, Visualization, Validation, Conceptualization. \textbf{Jiale Liu:} Writing - review \& editing, Writing - original draft, Visualization, Validation, Conceptualization. \textbf{Yu-Wei Zhan:} Writing - review \& editing, Conceptualization. \textbf{Zhen-Duo Chen:} Writing - review \& editing, Conceptualization. \textbf{Xin-Shun Xu:} Writing - review \& editing, Conceptualization.

\section*{Declaration of competing interest}
The authors declare that they have no known competing financial interests or personal relationships that could have appeared to influence the work reported in this paper.


\printcredits

\bibliographystyle{cas-model2-names}

\bibliography{sample-base}

\end{document}